\documentclass[letterpaper, 10 pt, conference]{ieeeconf}  

\IEEEoverridecommandlockouts                              

\usepackage[T1]{fontenc}
\usepackage{aecompl}

\usepackage{cite,soul}
\usepackage{amsmath,amssymb,amsfonts}
\usepackage{graphicx}
\usepackage{subfigure}
\usepackage{textcomp}
\usepackage{xcolor}
\usepackage{booktabs}
\usepackage{multirow}
\usepackage{epstopdf}
\usepackage{soul}

\newtheorem{remark}{Remark}
\usepackage{url}

\usepackage{algorithm}
\usepackage{algpseudocode}
\usepackage{amsmath}
\def\BibTeX{{\rm B\kern-.05em{\sc i\kern-.025em b}\kern-.08em
    T\kern-.1667em\lower.7ex\hbox{E}\kern-.125emX}}
    
\title{\LARGE \bf
Integrated Intention Prediction and Decision-Making with Spectrum Attention Net and Proximal Policy Optimization 
}

\author{Xiao Zhou, Chengzhen Meng, Wenru Liu, Zengqi Peng, Ming Liu, and Jun Ma
\thanks{This work was supported in part by the National Natural Science Foundation of China under Grant 62303390; and in part by the Guangzhou-HKUST(GZ) Joint Funding Scheme under Grant 2024A03J0618. \textit{(Corresponding Author: Jun Ma.)}}
\thanks{Xiao Zhou, Chengzhen Meng, Wenru Liu, Zengqi Peng, and Ming Liu are with the Robotics and Autonomous Systems Thrust, The Hong Kong University of Science and Technology (Guangzhou), Guangzhou, China. (e-mail: xzhou910@connect.hkust-gz.edu.cn; cmeng403@connect.hkust-gz.edu.cn; wliu354@connect.hkust-gz.edu.cn; zpeng940@connect.hkust-gz.edu.cn; eelium@hkust-gz.edu.cn).}
\thanks{Jun Ma is with the Robotics and Autonomous Systems Thrust, The Hong Kong University of Science and Technology (Guangzhou), Guangzhou, China, and also with the Division of Emerging Interdisciplinary Areas, The Hong Kong University of Science and Technology, Hong Kong SAR, China. (e-mail: jun.ma@ust.hk).} 
}

\begin{document}

\maketitle
\thispagestyle{empty}
\pagestyle{empty}

\begin{abstract}
For autonomous driving in highly dynamic environments, it is anticipated to predict the future behaviors of surrounding vehicles (SVs) and make safe and effective decisions.
However, modeling the inherent coupling effect between the prediction and decision-making modules has been a long-standing challenge, especially when there is a need to maintain appropriate computational efficiency.
To tackle these problems, we propose a novel integrated intention prediction and decision-making approach, which explicitly models the coupling relationship and achieves efficient computation.
Specifically, a spectrum attention net is designed to predict the intentions of SVs by capturing the trends of each frequency component over time and their interrelations.
Fast computation of the intention prediction module is attained as the predicted intentions are not decoded to trajectories in the executing process.
Furthermore, the proximal policy optimization (PPO) algorithm is employed to address the non-stationary problem in the framework through a modest policy update enabled by a clipping mechanism within its objective function. On the basis of these developments, the intention prediction and decision-making modules are integrated through joint learning.
Experiments are conducted in representative traffic scenarios, and the results reveal that the proposed integrated framework demonstrates superior performance over several deep reinforcement learning (DRL) baselines in terms of success rate, efficiency, and safety in driving tasks.

\end{abstract}

\section{INTRODUCTION}
Advancements in computer technology and artificial intelligence have led to significant developments in the field of autonomous driving technology. Nevertheless, the realization of autonomous driving systems remains a formidable challenge, with system design standing as a principal barrier. The architectures of contemporary autonomous driving systems fall into two major categories: end-to-end and hierarchical systems. While there have been notable advancements in end-to-end autonomous driving systems based on machine learning \cite{RN1}, these systems continue to grapple with issues of poor interpretability, high data demands, limited generalization capabilities, and insufficient robustness. Alternatively, hierarchical autonomous driving systems \cite{RN18, RN20}, which involve the process of perception, prediction, decision-making, planning, and control have exhibited considerable enhancements in terms of interpretability and robustness. 
However, for such systems, tasks including prediction and decision-making are often treated as relatively distinct modules, which ignores the inherent coupling relationship between them. This mutual effect comes from the influence of the prediction module on the decision-making module; and also, due to the interaction between vehicles, the decision-making module will in turn affect the behavior of the surrounding vehicles (SVs), thereby affecting the next prediction step. Since the coupling between prediction and decision-making could be overlooked, this separation can lead to deteriorated performance of autonomous vehicles (AVs) in complex situations.
\textcolor{black}{In response to this shortcoming, several studies have proposed frameworks that integrate the prediction and decision-making modules \cite{RN25,RN26}, leveraging predicted trajectories or intention to inform the decision-making modules in autonomous driving. In integrated frameworks such as \cite{huang2023differentiable}, the forecasting and imitation nets are trained jointly, which is more efficient compared to training the network individually.} Despite these efforts, the design of an integrated framework remains a significant challenge.

\textcolor{black}
{In general,} the output of the decision-making module can be obtained through approaches such as rule-based methods \cite{RN14}, optimization-based methods \cite{ma2022alternating}, and learning-based methods \cite{RN17}. 
Rule-based and optimization-based methods have obtained commendable results owing to their strong interpretability and related attributes. Nevertheless, due to the limited adaptability of these methods, they generally deliver poor performance in complex real-world environments. On the other hand, learning-based methods have demonstrated significant and efficacious progress, with a particularly promising direction in reinforcement learning (RL). A deep reinforcement learning (DRL) method presented in \cite{peng2023curriculum} adjusts the clipping parameter at different stages of training using proximal policy optimization (PPO), so that the vehicle could quickly search for an approximate optimal policy or its neighborhood with a large parameter and then converge to the optimal policy with a smaller one. The close interplay between prediction and decision-making implies that the prediction module exerts a considerable influence on decision-making module. Therefore, it is crucial to incorporate a well-developed prediction module in the framework to ensure that the decisions made by the system are safe and reliable.

Mainstream prediction methods typically focus on intention or trajectory prediction. Within the integrated framework, trajectory prediction is more commonly utilized and it can be divided into two main types: optimization-based and learning-based methods. Exemplified by \cite{RN2}, optimization-based methods are highly efficient in capturing long-term movements, but replicating this success in complex urban road environments proves challenging due to the intricate road layouts and variable traffic conditions. In contrast, learning-based methods \cite{RN3,RN4} are more adept at handling prediction tasks in such complex settings, benefiting from the strengths of neural networks in processing large amounts of trajectory data and understanding urban road conditions. Learning-based methods forecast future trajectories by analyzing historical trajectory data, resulting in encouraging outcomes. The performance can be improved when transforming trajectory representation from the time domain to the frequency domain, as the frequency domain reflects different characteristics such as energy distribution.
Relevant records have shown that these features in the frequency domain are critical to the prediction process. A novel approach for predicting state sequences through Fourier transform (FT) is proposed by \cite{NEURIPS2023_d5b94ca5}, which extracts hidden structural information from the frequency domain, thereby enhancing sample efficiency. A network called SpectrumNet is introduced in \cite{RN11} to predict pedestrian trajectories, which is capable of breaking down information across various time scales through FT and representing it in the frequency domain. In \cite{RN9}, the future trajectory is predicted from the history trajectory, which is transformed from the time domain to the frequency domain using FT. The low-frequency and high-frequency components reflect the long-term objective (LTO) and the short-term dynamic (STD) intention, respectively. However, the FT used in \cite{NEURIPS2023_d5b94ca5,RN11,RN9} fails to account for changes in frequency over time, hindering the dynamic estimation of trends in trajectory changes and affecting the performance of prediction in real-world scenarios. Given that the goal of trajectory prediction is to obtain precise future trajectories, it demands considerable computational resources, leading to difficulties in prediction within complex dynamic environments.
In addition to trajectory prediction, intention prediction has also been the subject of extensive research. For example, an intention prediction approach developed by \cite{RN6} utilizes Gaussian mixed hidden Markov models to forecast the short-term intention of vehicles based on their historical trajectories. However, this method concentrates exclusively on past trajectory data to predict short-term driving intention and does not take into account the LTO intention. The same issue can also be found in the majority of intention prediction approaches, such as  \cite{RN7}. The absence of consideration for LTO intention challenges the assurance of long-term decision-making effectiveness, which can compromise robustness in complex driving scenarios applied within integrated frameworks.

To address the aforementioned issues, we propose an integrated intention prediction and decision-making approach for autonomous driving, which leverages the proposed spectrum attention net to predict the intention of SVs at the frequency domain and utilizes the PPO algorithm to generate the decisions. The main contributions of this paper are as follows: 
An integrated intention prediction and decision-making framework is proposed to explicitly model the strong coupling effect between intention prediction and decision-making, where we use a joint learning mechanism to learn the coupling and ensure the effectiveness of network parameter updates.
Specifically, we propose the spectrum attention net to predict the intention of the SVs with the spectrum of trajectory obtained by short-time Fourier
transform (STFT). With the predicted intention of SVs (which is the combination of LTO and STD intention), the AV can make safer and more effective decisions.
The effectiveness of the proposed integrated framework is demonstrated by comparisons with existing methods in various driving scenarios, and the results indicate that the proposed approach outperforms other baseline methods in these driving tasks.

We organize the remainder of the article as follows: Section II introduces the preliminaries of the Markov decision process (MDP) and STFT, and also gives the problem statement. The details of the proposed integrated framework are given in Section III, including the design of the spectrum attention net, the PPO algorithm, and the joint learning mechanism in the framework.
Based on these findings, we conduct experiments in four different scenarios described in Section IV. Finally, we summarize the conclusion and future work in Section V.

\section{Preliminaries}
\subsection{Markov Decision Process}
MDP is used to describe decision-making problems with the Markov property. In an MDP, the agent selects the optimal decision based on the current state and possible actions to maximize long-term rewards. 
The MDP consists of five elements, which can be represented by the tuple $\langle \mathcal{S}, \mathcal{A}, \mathcal{P}, \mathcal{R}, \gamma \rangle$, where $\mathcal{S}$ is the state space, $\mathcal{A}$ is the action space, $\mathcal{P}$ is the state transition probability function that represents the probability of transitioning from a past state to a new state after taking the action. $\mathcal{R}$ is the reward function that represents the immediate reward obtained after taking action $a$ in state $s$, and $\gamma \in(0,1)$ is the discount factor used to discount the value of future rewards.

\subsection{Short-Time Fourier Transform}

FT is pivotal in applications that need to transform signals between the time and frequency domains. An important variant is the STFT \cite{allen1977unified}, which addresses the limitation of the FT in analyzing non-stationary signals where frequency components vary over time. The primary purpose of the STFT is to determine the frequency and phase content of local sections of a signal as it changes over time. This is achieved by multiplying the signal by a window function which is shifted along the duration of the signal. For a signal $x[n]$ where $n$ represents the time step, the STFT transforms it into the frequency domain by applying FT to segments of the signal, defined by the window function. The STFT can be expressed as:
\begin{equation}
\begin{split}
X(m, \omega) = \sum_{n=-\infty}^{\infty} x[n] \cdot w[n-m] \cdot e^{-j \omega n}
\end{split}
\label{state martix}
\end{equation}
where \(w[n-m]\) denotes the window function centered around time \(m\), $m$ is the index of time frame, and \(\omega\) corresponds to the frequency.

On the other hand, when the frequency domain signal obtained from STFT needs to be converted to the time domain, the inverse short-time Fourier transform (ISTFT) can be adopted.
Given the STFT representation $X(m, \omega)$ of a signal, the corresponding signal $x(n)$ at time domain  can be reconstructed using the formula:
\begin{equation}
x(n) = \sum_{m} \sum_{\omega} X(m, \omega) \cdot w(n-m) \cdot e^{j\omega n}
\end{equation}

\subsection{Problem Statement}

The objective of this research is to develop a framework for effectively navigating AV in various traffic environments. We will focus on four key scenarios: a straight roadway, a four-way intersection, a two-way intersection, and a roundabout.

By modeling this problem as an MDP, our goal is to find an optimal policy $\pi_{\theta}$, guided by parameters $\theta$, that maximizes expected cumulative discounted rewards, formulated as:
\begin{equation}
\begin{array}{cl}
\max _{\pi_{\theta}} & \mathbb{E}\left[\sum_{t=1}^T \gamma^t r_{t}\left(\mathbf{S}_t, a_{t}\right)\right] \\
\text { s.t. } & a_{ t} \in \mathcal{A}, \mathbf{S}_t \in \mathcal{S}
\end{array}
\end{equation}

Specifically, the state space $\mathcal{S}$ and action space $\mathcal{A}$ in the problem is defined as:

\textbf{State space $\mathcal{S}$}: We define the state at any given time as a matrix, 
\begin{equation}
\begin{split}
\mathbf{S}_t=\left[\ \left[\mathbf{s}_t^0\right]^T \ \ \left[\mathbf{s}_t^1\right]^T \ ...\ \left[\mathbf{s}_t^N\  \right]^T \right]^T
\end{split}
\label{state martix}
\end{equation}
where $N$ is the number of vehicles. The state of the AV is captured in the first row of $\mathbf{S}_t$ as $\mathbf{s}_t^0$, and other rows of $\mathbf{S}_t$, i.e., $\mathbf{s}_t^i\ (i=1,2,...,N)$, represent vectors of the kinematic features of the SVs. Each vehicle's state is characterized by the following kinematic features: 
\begin{equation}
\begin{split}
\mathbf{s}_t^{i}={\left[\begin{array}{l l l l l l l}{x_t^{i}}&{y_t^{i}}&{v_{x,t}^{i}}&{v_{y,t}^{i}}&{\sin\psi_t^{i}}&{\cos\psi_t^{i}}\end{array}\right]}
\end{split}
\label{kinematic_features}
\end{equation}
which contains the information about the position, velocity, and orientation of vehicle $i$ at time step $t$.

\textbf{Action space $\mathcal{A}$}: The agent has access to a set of five discrete actions, 
\begin{equation}
\begin{split}
\mathcal{A}=\left\{ A^{0},A^{1},A^{2},A^{3},A^{4} \right\}
\end{split}
\label{Action space}
\end{equation}
including lateral maneuvers such as changing lanes to the left or right ($A^{0}$, $A^{2}$), maintaining current motion ($A^{1}$), and longitudinal adjustments like decelerating or accelerating ($A^{3}$, $A^{4}$).





\section{METHODOLOGY}
\begin{figure*}[!t] 
    \centering   
    \includegraphics[trim=0.3cm 0 0 0, width=0.95\linewidth]{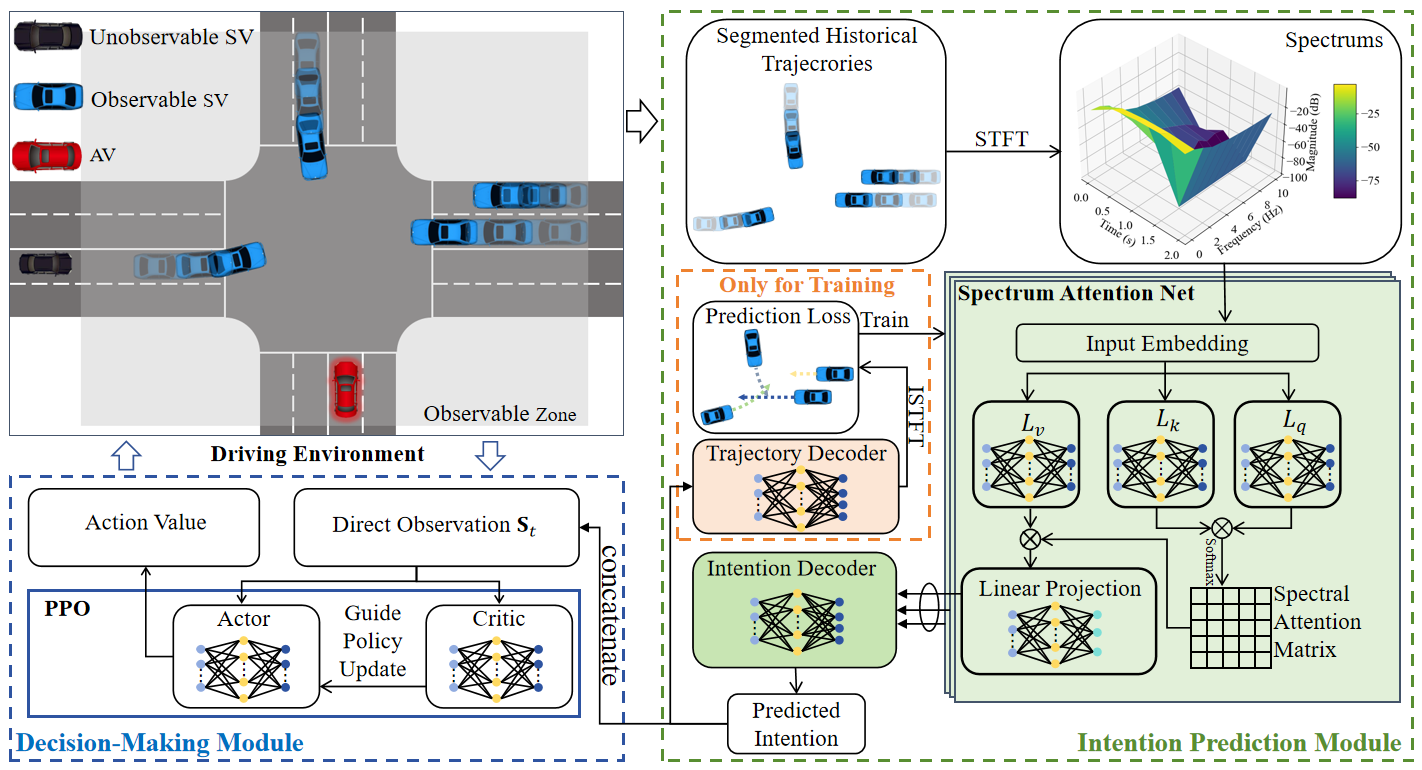}
   \caption{Overall architecture of the proposed integrated intention prediction and decision-making framework is exemplified through a four-way intersection scenario. For observable SV $i$, its segmented historical trajectory is transformed into a spectrogram by STFT first, then used to predict its intention with the spectrum attention net. The decision-making module receives the predicted intention of SVs and makes a decision based on predicted intention and direct observation. Note that the trajectory decoder is only executed in the joint training process to improve the real-time capabilities of the integrated framework.  
   }
    \label{Net_frame}
\end{figure*}

In this section, we present an inaugural framework for integrated intention prediction and decision-making as shown in Fig. \ref{Net_frame}, where the AV utilizes the intention prediction obtained from the spectrum attention net to guide its decisions.
We first introduce the details of the proposed spectrum attention net, which is used to predict the intention of SVs. 
Then, we illustrate the advantages of the PPO algorithm in processing complex 
and non-stationary inputs as a decision-making module in the integrated framework.
Finally, we give the details about the joint learning process of the intention prediction module and decision-making module in the integrated framework.

\subsection{Intention Prediction Module with Spectrum Attention Net}

The behavior of SVs is affected by both LTO and STD intention, which can be reflected by the low-frequency component and high-frequency component in the spectrum, respectively. Therefore, we propose a spectrum attention net to predict the intention of SVs from the frequency spectrum, where the spectral information can characterize the intention of SVs.

At time step $t$, the AV observes current environmental states, which include information about both itself and observable SVs. The state information of these observable SVs is recorded in the trajectory buffer with a horizon of $H$ to maintain a consistent input length for the spectrum attention net. Specifically, for each observable vehicle $i$ in the scenario, its state information $\mathbf{s}^i_t$ over the time interval $[t-H+1, t]$ is incorporated into the trajectory buffer $T_{ego, i}$, thereby forming a segmented historical trajectory for vehicle $i$. 
By utilizing the segmented historical trajectory as time sequence input, we derive the features of multiple frequency components evolving over time through the transformation of time sequence input into the frequency domain using STFT, expressed as follows:
\begin{equation}
\begin{split}
X_s(m, \omega) = \sum_{n=t-H+1}^{t} \mathbf{s}^i_t \cdot w[t-m] \cdot e^{-j \omega t}
\end{split}
\label{state martix}
\end{equation}
where $X_s(m, \omega) \in \mathbb{R}^{F \times \Omega \times M}$ denotes the trend of the frequency component at $\omega$ across time frame $m$. Note that $\Omega$ is the number of frequency components, $M$ is the number of time frames, and $F$ denotes the dimension of kinematic features.
Indeed, STFT can be regarded as a feature extraction function that derives intent representation from a segmented historical trajectory. The fixed time-frequency resolution of STFT ensures uniform representation of each time frame, which is crucial for the attention mechanism to effectively discern and correlate different temporal segments, thereby facilitating accurate intention predictions.

The behavior of SVs involves weighing different levels of intention, and the intention is interrelated. The frequency components at different time frames within the spectrogram $X_s(m, \omega)$ depict the distribution of intention across time.
Therefore, to thoroughly research intention patterns across diverse combinations of frequency components, we introduce a multi-head spectrum attention net in this study to forecast the intention of SVs.
Initially, the spectrogram {$X_s(m, \omega)$} is embedded with a linear projection and then sent to the spectral attention module.
Through the transformations executed by networks $L_q$, $L_k$, and $L_v$, all embedded vectors are converted into query $Q\in \mathbb{R}^{F \times \Omega \times M}$, key $K\in \mathbb{R}^{F \times \Omega \times M}$, and value $V\in \mathbb{R}^{F \times \Omega \times M}$, respectively. The attention scores are computed by multiplying query matrix $Q$ with the transpose of $K$, followed by scaling with the inverse square root of dimension $M$ and normalization via a softmax function across the frequency dimension. 
\textcolor{black}{This process yields a spectral attention matrix, which is utilized to evaluate the correlation among different frequency components, thereby facilitating the prediction about the intention of vehicle $i$}. The resulting prediction vector is subsequently fed into the linear projection network, with each head of spectral attention generating outputs as represented below:

\begin{equation}
P_t^i = \text{softmax}\left(\frac{Q K^{T}}{\sqrt{d_k}}\right) V
\label{Equ::intention prediction}
\end{equation}
Finally, the outputs from all heads are combined and decoded by the intention decoder, ultimately serving as the predicted intention of SV.

\begin{remark}
The behavior of SVs is affected by multiple levels of intention at the same time, including LTO and STD intention. For a spectrogram $X_s(m, \omega)$ that delineates frequency domain variations over time, the attention mechanism can capture the trend of each frequency component over time and their interrelations. Therefore, reasonable predictions about the spectrum distribution in a future time horizon can be obtained by spectrum attention net and reflect the intention of SVs.
\end{remark}
\subsection{Decision-Making Module with PPO}
One of the main challenges in the decision-making module of the integrated framework is the non-stationary problem led by joint learning. In the joint learning process, the intention prediction in the observation input of the decision-making algorithm keeps changing, therefore the decision-making policy needs to be adjusted accordingly. As the spectrum attention net in the intention prediction module updates, corresponding adjustments of the policy in the decision-making module become necessary. \textcolor{black}{However, unchecked policy updates can lead to catastrophic consequences in this scenario. Significant updates on the policy of the decision-making module based on current intent prediction module may lead to excellent performance in current environment, but the performance could be suboptimal or even terrible after the intent prediction module updates its network parameters.} To address this concern, we adopt the PPO algorithm in the integrated framework. 

\textcolor{black}{The PPO algorithm is an online reinforcement learning framework designed to solve sequential decision-making problems.
Notably, the PPO algorithm employs a clipping mechanism within its objective function to ensure modest policy updates.} This feature becomes especially crucial when handling complex and non-stationary multimodal inputs from the intention prediction module.
Specifically, the objective function of PPO designed to mitigate the risk of destabilizing updates is given by:

\begin{equation}
L^{\text{CLIP}}(\theta) = \hat{\mathbb{E}}_t \left[ \min \left( r_t(\theta) \hat{A}_t, \text{clip}\left(r_t(\theta), 1-\epsilon, 1+\epsilon\right) \hat{A}_t \right) \right]
\label{equ::Lclip}
\end{equation}
where \( r_t(\theta) = \frac{\pi_\theta(a_t | s_t)}{\pi_{\theta_{old}}(a_t | s_t)} \) represents the probability ratio of selecting action \( a_t \) in state \( s_t \) under the new policy \( \pi_\theta \) compared to the old policy \( \pi_{\theta_{old}} \). Additionally, \( \hat{A}_t \) denotes the estimated advantage function at timestep \( t \), and \( \epsilon \) serves as a hyperparameter defining the clipping boundary.

Essentially, the clipped surrogate objective in PPO provides a theoretically constrained network update routine to tackle the non-stationary problem associated with the intention prediction module. This mechanism ensures a stable policy update step within the clipping boundary, which is indispensable for the algorithm's convergence and stationarity in a non-stationary environment.

\subsection{Joint Learning Process}
The integrated intention prediction and decision-making framework is proposed in this study utilizing the spectrum attention net and PPO algorithm. The joint training of the intention prediction and decision-making modules is employed, as it outperforms the standalone training mechanism~\cite{huang2023differentiable}.

For the joint learning process in the integrated framework, the agent needs to explore the environment and collect experience based on the current network parameters first. In detail, the spectrum attention net predicts SVs' intention and the PPO algorithm makes decisions based on the observation and predicted intention to explore the environment and collect experiences. Then, the collected experiences serve as training data for the spectrum attention net and the PPO algorithm in the integrated framework. Specifically, the predicted intention recorded in the experiences is sent to the trajectory decoder and transformed to trajectory time sequences in the future time horizon $[t+1,t+H]$ through ISTFT transformation. Subsequently, the trajectory time sequences are compared with the ground truth recorded in the collected experience to calculate the prediction loss. Finally, the spectrum attention net is optimized by minimizing the prediction loss with stochastic gradient descent.
Meanwhile, the collected experiences contribute to optimizing the network parameters in the PPO algorithm by maximizing the objective function as in (\ref{equ::Lclip}).
To sum up, the learning process in the integrated framework is outlined in Algorithm \ref{I2PDM Alg}.


\begin{algorithm}
    \caption{Integrated intention prediction and decision-making algorithm}  \label{I2PDM Alg}
    \begin{algorithmic}[1]
        \State Initialize spectrum attention net with parameters $\omega$, policy (actor) network with parameters $\theta$ and value (critic) network with parameters $\phi$.
        \State Initialize trajectory buffer $B_t$ to capacity $H$, training epochs per collect $E$, trajectory length $T$, batch size $B$.
        \For{$k = 0, 1, 2, \dots$}
            \State Collect set of experiences $\mathcal{D}_k = \{\tau_i\}$ by running policy $\pi_{\theta_k}$ interaction with environment.
             \For{time  $t=0, 1, \ldots , T$}
                \For{observable vehicle $i=1, 2, \ldots , N$}
                    \State Record state of vehicle $i$ in $B_t^i$, transform the segmented historical trajectory into the frequency domain with STFT
                    \State Predict the intention of observable vehicle $P_t^i$ by spectrum attention net based on $B_t^i$
                \EndFor
                \State Update the predicted intention of observable SVs $P_t = \{P_t^1, P_t^2, \ldots, P_t^N\}$
                \State Select action by running policy $\pi_{\theta_k}$
                \State Execute action $a_t$ and observe reward $r_t$ and new state $\mathbf{S}_{t+1}$
                \State Store transition $\left(\mathbf{S}_{t}, P_t, \mathbf{a}_t, r_t, \mathbf{S}_{ t+1}\right)$ 
            \EndFor
            \State Compute trajectory target return estimates $\hat{R}_t$.
            \State Compute advantage estimates $\hat{A}_k$ with value $\hat{V}_{\phi}$.
            \For{$e = 0, 1, \dots, E-1$}
                \For{minibatch $b \in \mathcal{D}_k$}
                    \State Decode the predicted intention $P_t$ with the trajectory decoder and transform it into the trajectory time sequence by ISTFT
                    \State Update the  spectrum attention net by minimizing the prediction loss with SGD
                    \State $L_{pred}=\frac{1}{H \cdot F} \sum_{\tau=t+1}^{t+H}\left\|\hat{\mathbf{s}}_i^\tau-\mathbf{s}_i^\tau\right\|_2$
                    \State Update the policy by maximizing the PPO-Clip objective as in (\ref{equ::Lclip}) with SGD 
                \EndFor
                \State Fit the value by regression on mean-squared error with SGD:
                \State $L^{\text{VF}}(\phi) = \frac{1}{B \cdot T} \sum_{\tau \in b} \sum_{t=0}^{T-1} (V_{\phi}(s_t) - \hat{R}_t)^2$
            \EndFor
        \EndFor
    \end{algorithmic}
\end{algorithm}

\begin{remark}
The prediction and decision-making modules of autonomous driving systems are highly coupled in the conventional setting. Therefore, in this work, we explicitly model the coupling relationship through joint learning of the prediction and the decision-making modules. Specifically, in the proposed integrated framework, the spectrum attention net and the decision-making algorithm use the identical data for training and then update parameters simultaneously. 
It is worth noting that unlike end-to-end methods which optimize all networks with decision-making objective function, we optimize the spectrum attention net through the prediction loss, thus ensuring the effectiveness of network parameter updates.
\end{remark}

\begin{figure*}[htbp]
\centering
\label{reward curv}
\subfigure[Straight Road]{
\begin{minipage}[t]{0.24\linewidth}
\includegraphics[height=3.7cm]{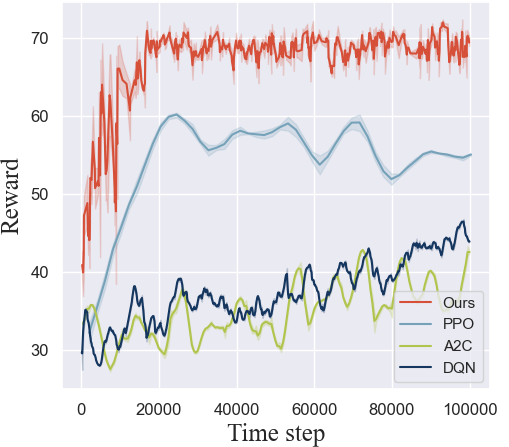}
\end{minipage}%
}%
\subfigure[Intersection-v0]{
\begin{minipage}[t]{0.24\linewidth}
\centering
\includegraphics[height=3.7cm]{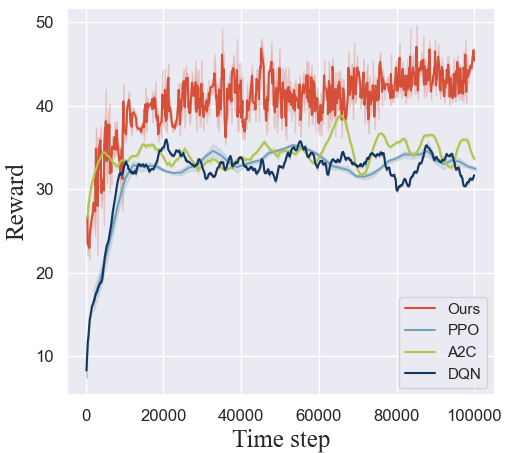}
\end{minipage}%
}%
\centering
\subfigure[Intersection-v1]{
\begin{minipage}[t]{0.24\linewidth}
\centering
\includegraphics[height=3.7cm]{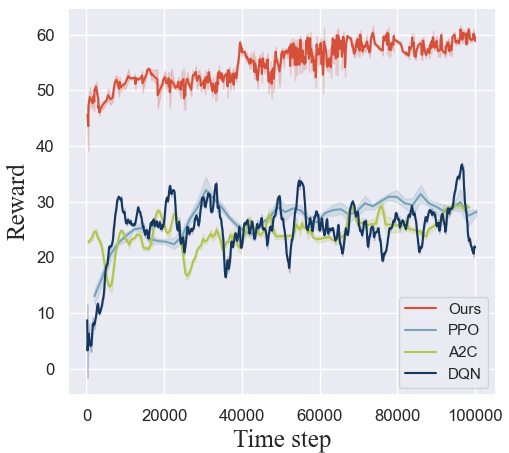}
\end{minipage}%
}
\subfigure[Roundabout]{
\begin{minipage}[t]{0.24\linewidth}
\centering
\includegraphics[height=3.7cm]{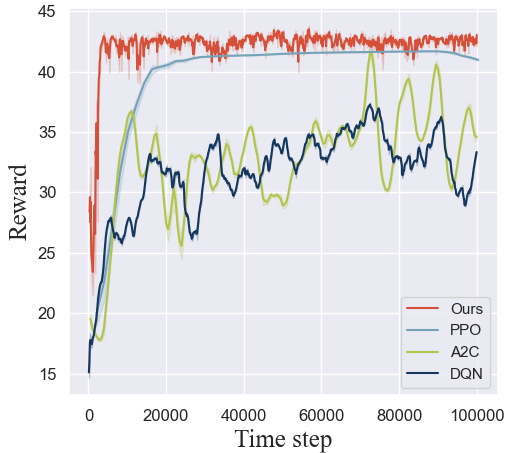}
\end{minipage}%
}
\caption{Learning curves of different methods in four representative scenarios. The training curves are smoothed by the Savitzky-Golay filter.}
\label{scenrios}
\end{figure*}

\section{EXPERIMENTS}

In this section, we test the proposed integrated framework in the four common autonomous driving scenarios and compare the results with other classic DRL algorithms. The simulations are conducted in an OpenAI Gym environment $Highway\_Env$ \cite{highway-env}, and the behaviors of SVs are characterized by the intelligent driver model (IDM) \cite{treiber2000congested}. and the baseline DRL methods are implemented based on Stable-baselines3 \cite{stable-baselines3}.



\begin{table}[t]
\centering
\caption{Hyperparameters}
\label{tab:hyperparameters}
\begin{tabular}{@{}l@{\hspace{8pt}}c@{\hspace{8pt}}c@{\hspace{8pt}}c@{\hspace{8pt}}c@{}}
\toprule
Hyperparameter          & DQN     & A2C     & PPO   & Ours  \\ \midrule
Discount Factor         & $0.99$ & $0.99$ & $0.99$ & $0.99$ \\
Learning Rate           & $0.0005$  & $0.0005$  & $0.0005$  & $0.0005$\\
Batch Size              & $64$ & $64$ & $64$  & $64$\\
Exploration Rate        & $0.1$   & -         & -  & -\\
Target Network Update Frequency   & $5000$    & -         & -  & -\\
Entropy Coefficient     & -   & $0.0$   & $0.0$   & $0.0$       \\
Value Function Coefficient & -    & $0.5$    & $0.5$   & $0.5$       \\
Number of Steps         & - & $1024$ & $1024$  & $1024$        \\
\bottomrule
\end{tabular}
\end{table}

\subsection{Experimental Setup}


\textbf{Straight Road}: In this scenario, the AV drives on a four-lane straight road filled with SVs, with the task of safely driving 600 meters without collisions within 1 minute. The reward function is defined as $r = r_{\text {Rel}}+r_{\text{R}}+r_{\text {col}}+r_{\text{LC}}+r_{\text{speed}}$, where $r_{\text {Rel}}$ and $r_{\text{R}}$ represent the rewards of completing the task and driving on the right side of the road. $r_{\text {col}}$ and $r_{\text{LC}}$ are the penalties of collision and lane change, which can reduce unnecessary lane-changing behavior. $r_{\text{speed}}$ will reward high-speed driving behavior within the speed limitation while penalizing for exceeding the speed limitation.

\textbf{Intersection-v0}: This scenario is a four-way intersection, with the task of safely and quickly driving toward the designated lane. The reward function is settled as $r = r_{\text {Rel}}+r_{\text {col}}\left(v, N\right) +r_{\text{OfR}}+r_{\text{speed}}$, where $r_{\text{OfR}}$ is the penalty of out of the road and $\left(v, N\right)$ is the speed of the AV and number of SVs in every time interval, which reflects the collision rate.

\textbf{Intersection-v1}: It is similar to Intersection-v0, except that there are two lanes in one direction, and vehicles are allowed to change lanes. Therefore, a lane-changing penalty is added to the reward function, which is written as $r = r_{\text {Rel}}+r_{\text {col}}\left(v, N\right) +r_\text{OfR}+r_{\text{LC}}+r_{\text{speed}}$.

\textbf{Roundabout}: In this scenario, there is a roundabout with two lanes, with the task of passing through the roundabout as soon as possible without collision. To prevent vehicles from circling, a penalty  of timeout $r_{\text{TO}}$ is added in the reward function, which can be expressed as $r = r_{\text {Rel}}+r_{\text {col}}\left(v, N\right) +r_{\text{OfR}}+r_{\text{LC}}+r_{\text{TO}}+r_{\text{speed}}$

 To validate the effectiveness of the proposed method, we evaluate its performance in the above scenarios and compare it with other three baseline DRL methods. The number of SVs is 8. In each simulation, we randomize the states of all agents to prevent the policy network from memorizing actions. We compare the proposed framework with the following baseline methods:

\begin{itemize}
\item A2C performs gradient ascent on the parameter $\theta$ by $\nabla_\theta J(\theta)=\mathbb{E}_\tau\left[\sum_{t=0}^{T-1} \nabla_\theta \log \pi_\theta\left(a_t \mid s_t\right) G_t\right]$ to directly maximize the reward \cite{mnih2016asynchronous}. 
\item DQN updates the Q function using the SGD algorithm by minimizing the loss $\mathcal{L}=1 / N \sum_{j=0}^{N-1}\left(Q\left(s_j, a_j\right)-y_j\right)^2$ \cite{mnih2013playing}.
\item PPO updates indirectly by maximizing a surrogate objective function as in (\ref{equ::Lclip}), which provides a cautious estimate of the potential change in $J(\pi_{\theta})$ resulting from the update \cite{schulman2017proximal}. This method is used as an ablation study.
\end{itemize}

The hyperparameters for network training of the four algorithms are listed in Table I, and all networks use Adam as the optimizer. The experiments with the aforementioned scenarios are conducted on the Windows 11 system environment with an Intel Core i7-14700KF processor and an NVIDIA RTX 4070 Ti GPU.

\subsection{Performance Demonstration}

\begin{figure*}[htbp]
\centering
\subfigure[Time step $t=60$]{
\begin{minipage}[t]{0.22\linewidth}
\centering
\includegraphics[width=0.9\linewidth]{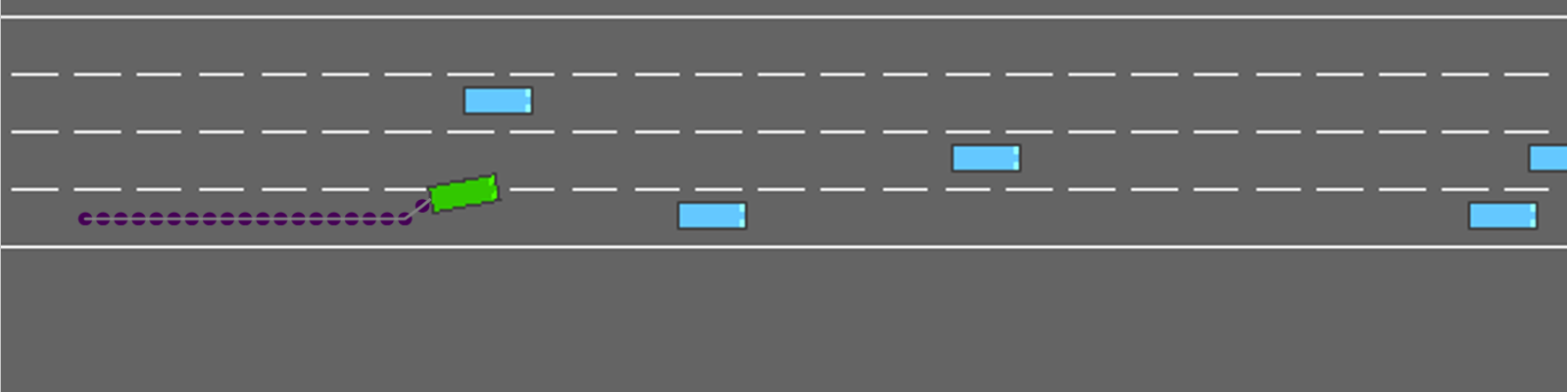}
\end{minipage}%
}%
\subfigure[Time step $t=100$]{
\begin{minipage}[t]{0.22\linewidth}
\centering
\includegraphics[width=0.9\linewidth]{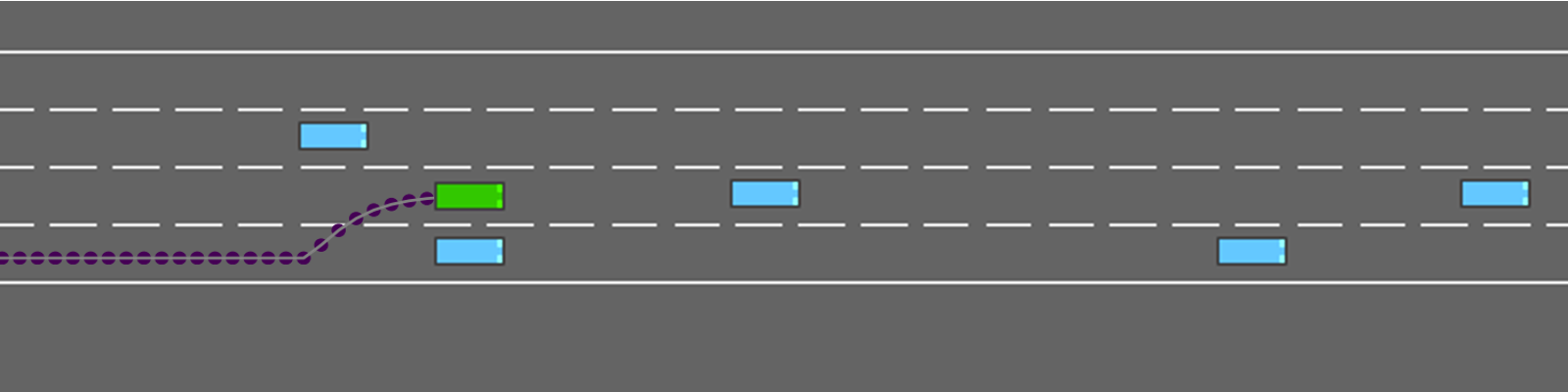}
\end{minipage}%
}%
\centering
\subfigure[Time step $t=115$]{
\begin{minipage}[t]{0.22\linewidth}
\centering
\includegraphics[width=0.9\linewidth]{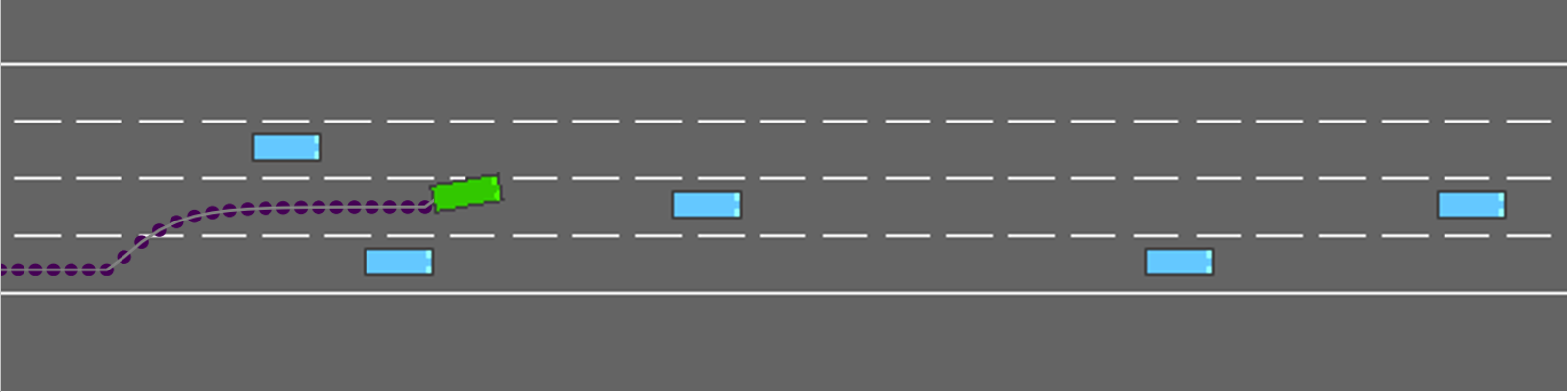}
\end{minipage}%
}
\subfigure[Time step $t=140$]{
\begin{minipage}[t]{0.22\linewidth}
\centering
\includegraphics[width=0.9\linewidth]{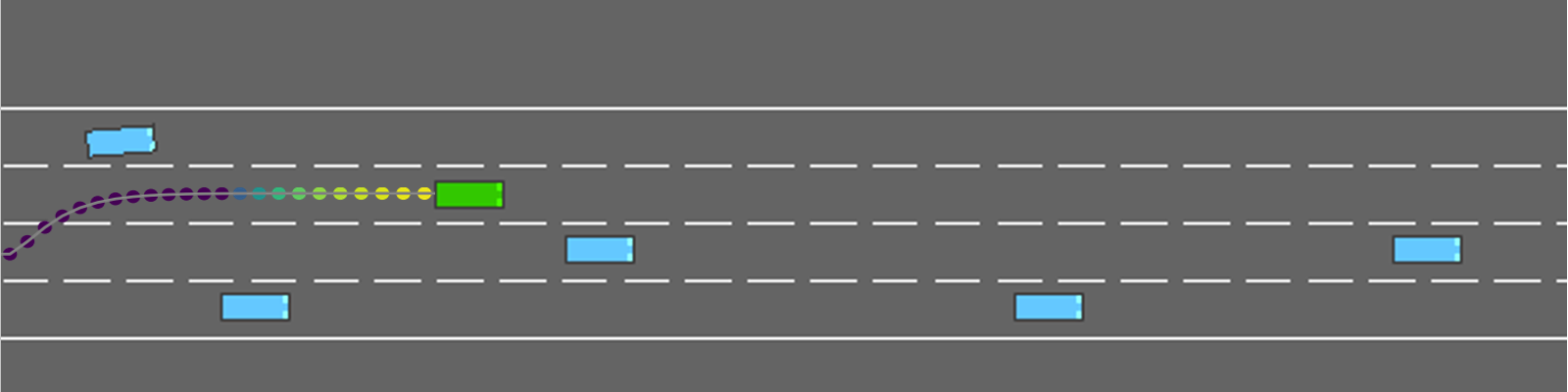}
\end{minipage}%
}

\subfigure[Time step $t=15$]{
\begin{minipage}[t]{0.22\linewidth}
\centering
\includegraphics[width=0.9\linewidth]{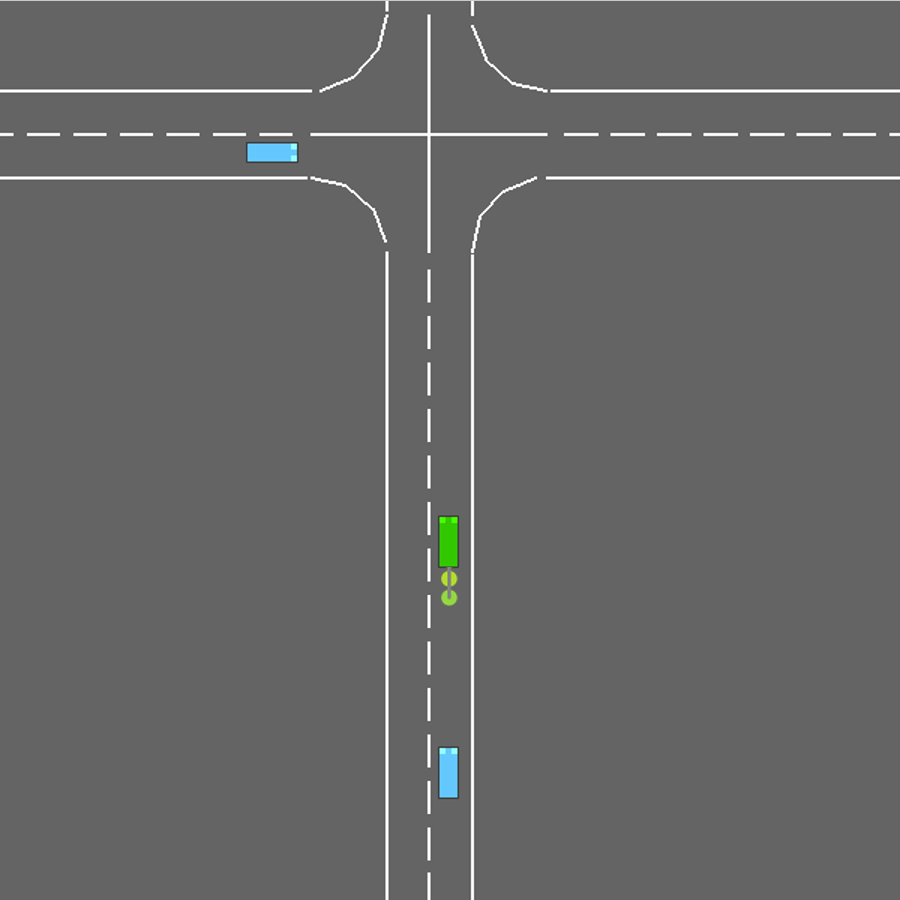}
\end{minipage}%
}%
\subfigure[Time step $t=35$]{
\begin{minipage}[t]{0.22\linewidth}
\centering
\includegraphics[width=0.9\linewidth]{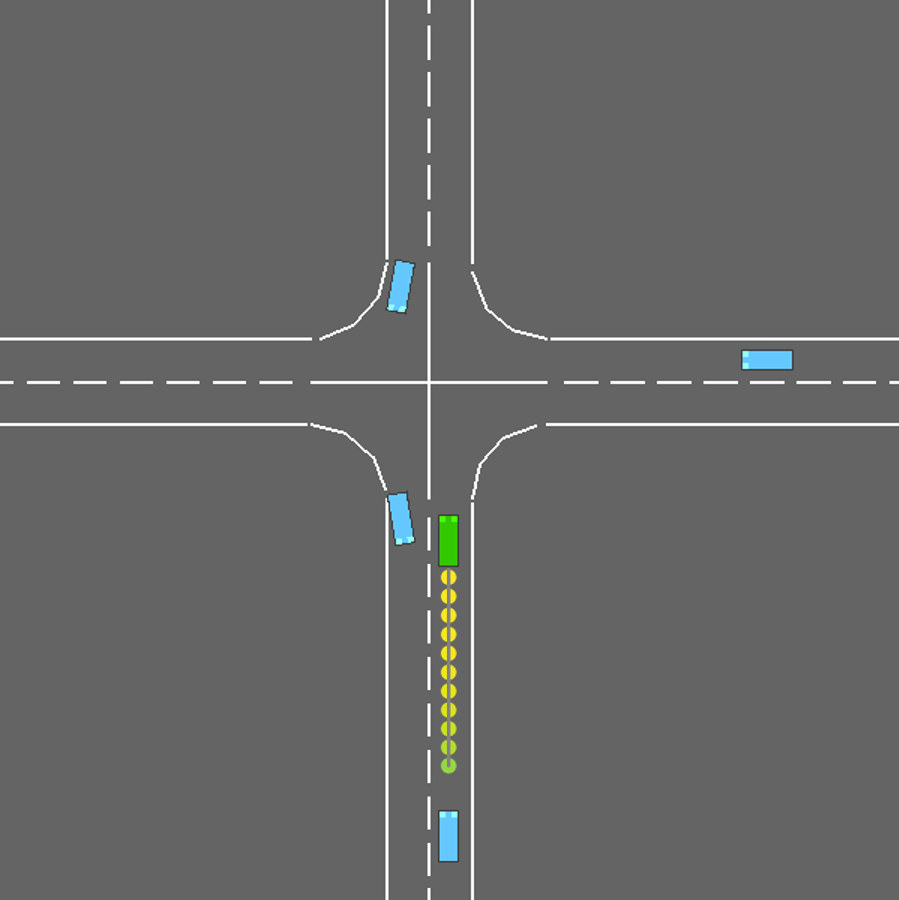}
\end{minipage}%
}%
\centering
\subfigure[Time step $t=65$]{
\begin{minipage}[t]{0.22\linewidth}
\centering
\includegraphics[width=0.9\linewidth]{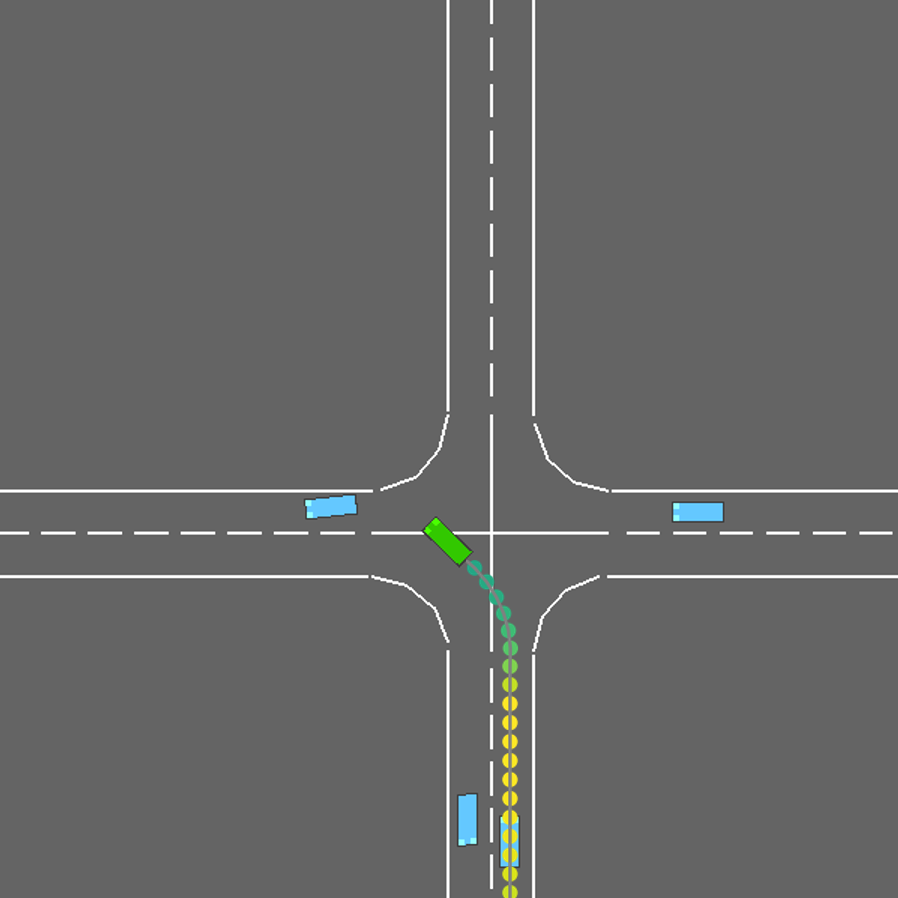}
\end{minipage}%
}
\subfigure[Time step $t=115$]{
\begin{minipage}[t]{0.22\linewidth}
\centering
\includegraphics[width=0.9\linewidth]{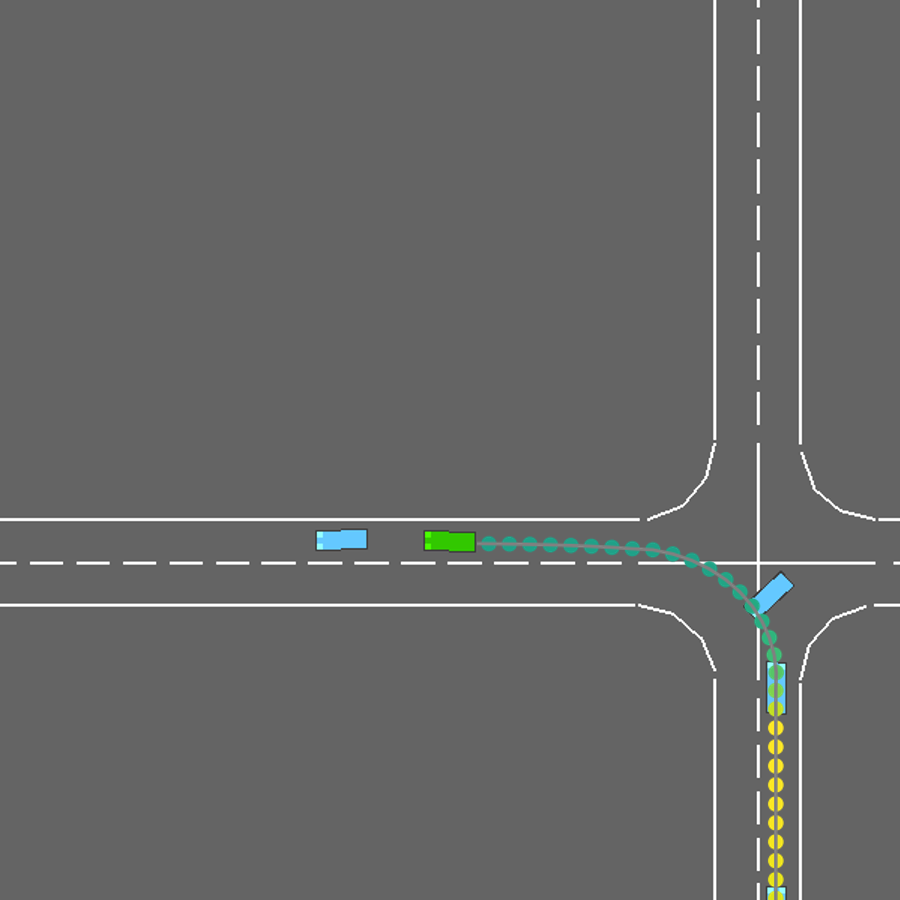}
\end{minipage}%
}

\subfigure[Time step $t=20$]{
\begin{minipage}[t]{0.22\linewidth}
\centering
\includegraphics[width=0.9\linewidth]{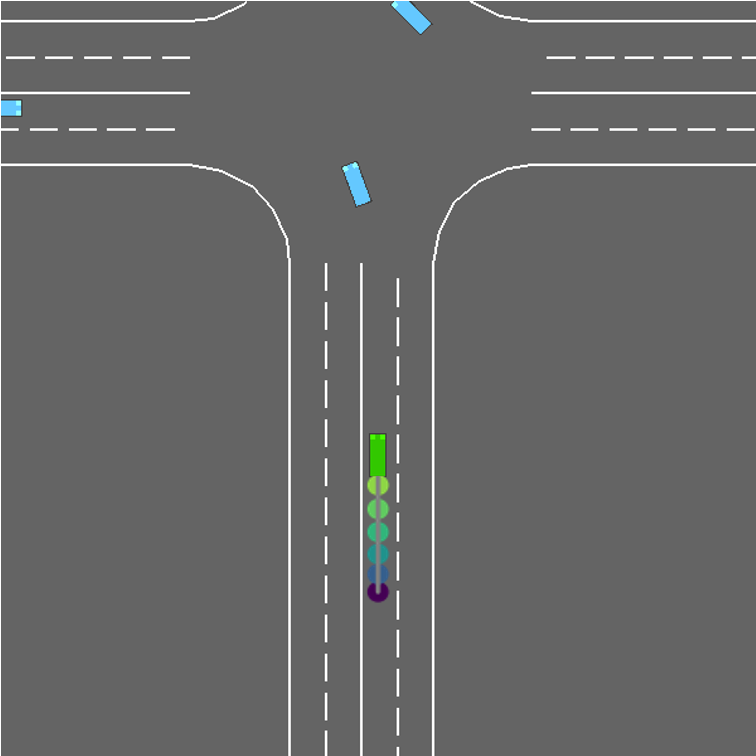}
\end{minipage}%
}%
\subfigure[Time step $t=40$]{
\begin{minipage}[t]{0.22\linewidth}
\centering
\includegraphics[width=0.9\linewidth]{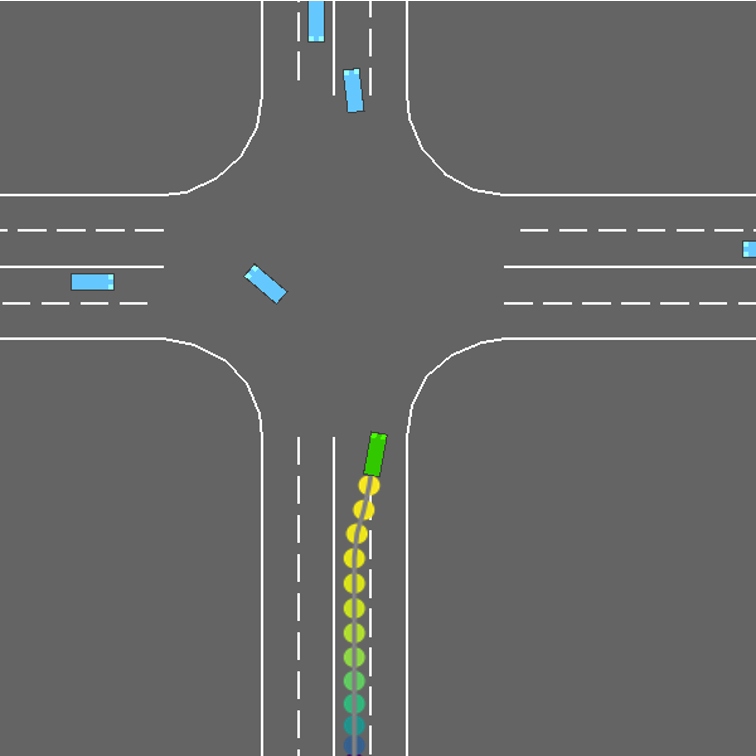}
\end{minipage}%
}%
\centering
\subfigure[Time step $t=65$]{
\begin{minipage}[t]{0.22\linewidth}
\centering
\includegraphics[width=0.9\linewidth]{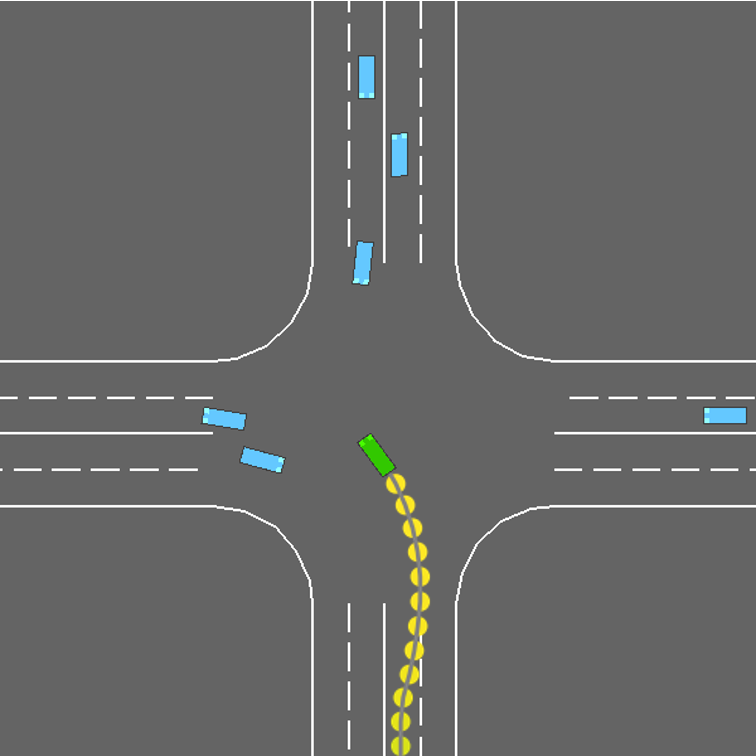}
\end{minipage}%
}
\subfigure[Time step $t=90$]{
\begin{minipage}[t]{0.22\linewidth}
\centering
\includegraphics[width=0.9\linewidth]{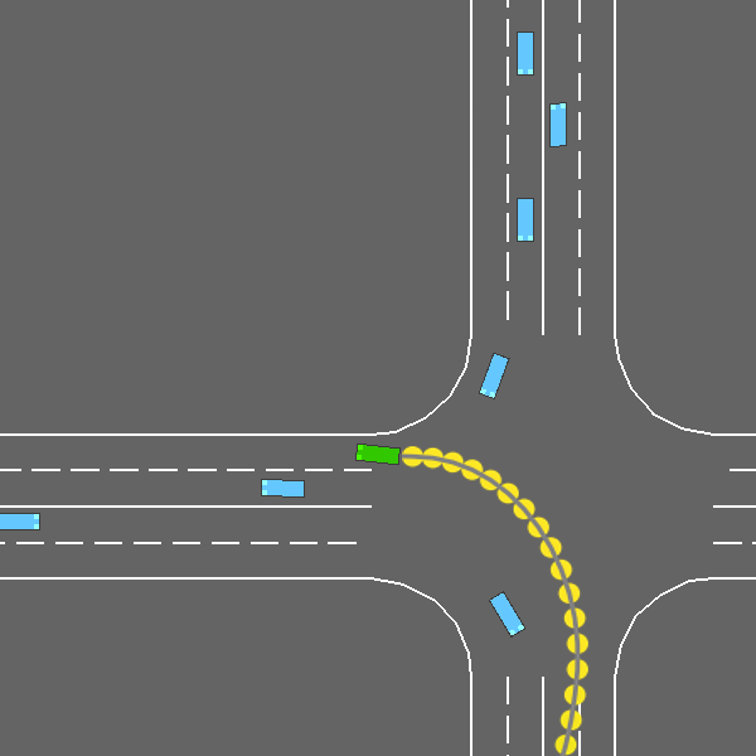}
\end{minipage}%
}

\subfigure[Time step $t=20$]{
\begin{minipage}[t]{0.22\linewidth}
\centering
\includegraphics[width=0.9\linewidth]{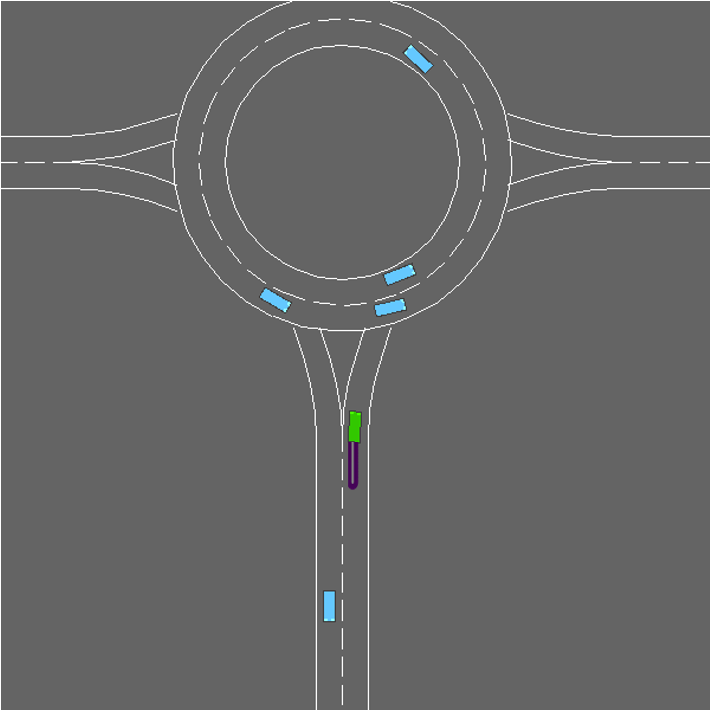}
\end{minipage}%
}%
\subfigure[Time step $t=60$]{
\begin{minipage}[t]{0.22\linewidth}
\centering
\includegraphics[width=0.9\linewidth]{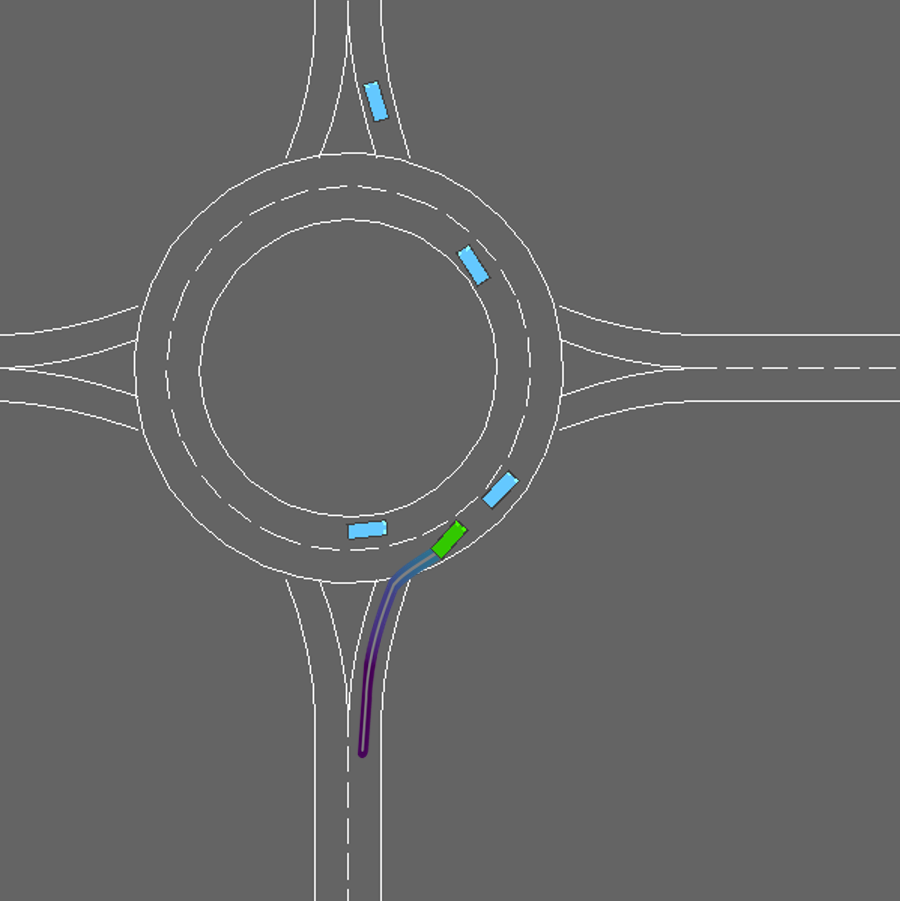}
\end{minipage}%
}%
\centering
\subfigure[Time step $t=110$]{
\begin{minipage}[t]{0.22\linewidth}
\centering
\includegraphics[width=0.9\linewidth]{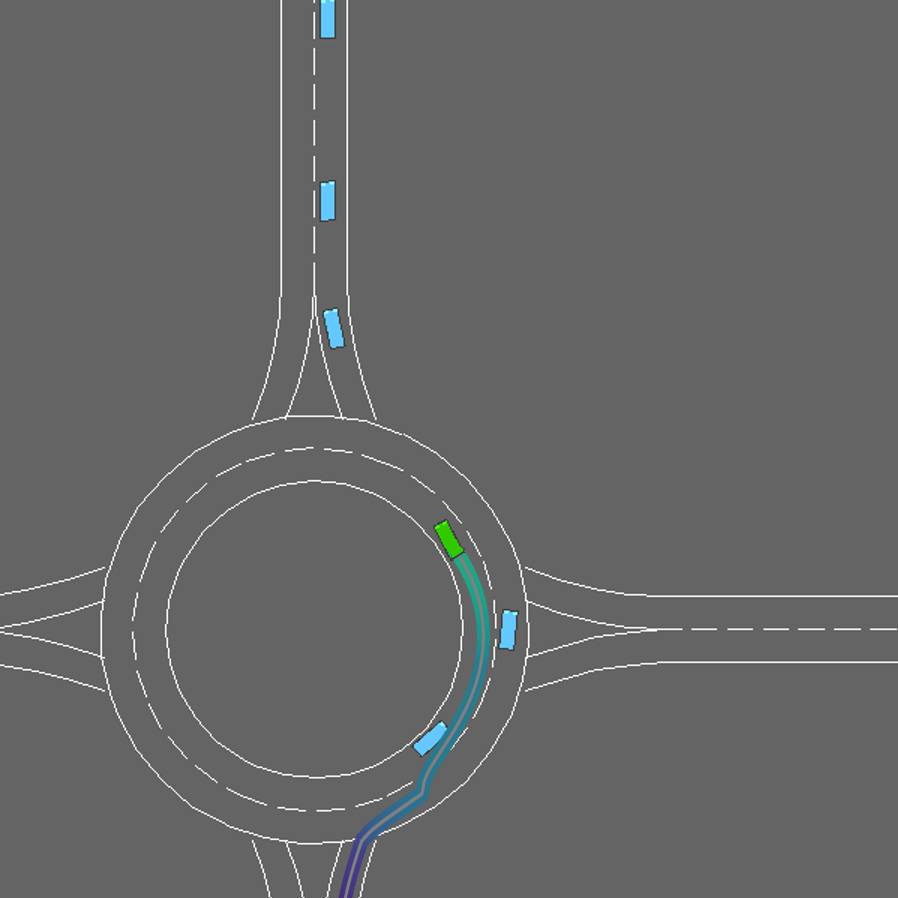}
\end{minipage}%
}
\subfigure[Time step $t=150$]{
\begin{minipage}[t]{0.22\linewidth}
\centering
\includegraphics[width=0.9\linewidth]{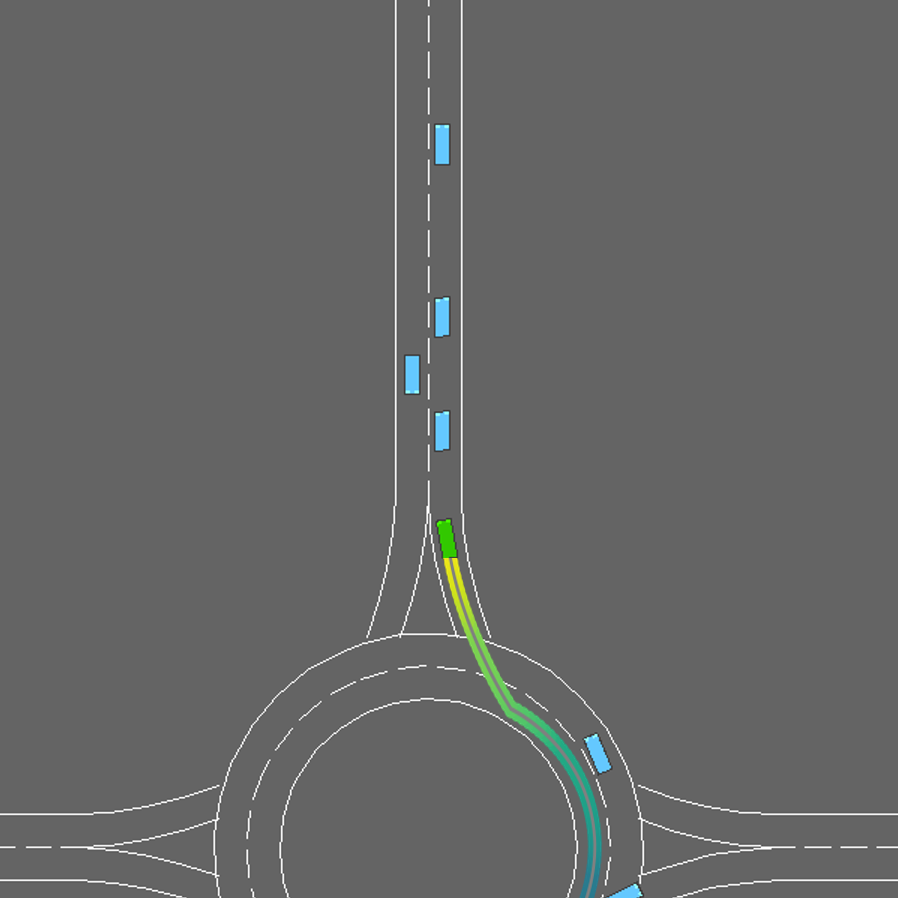}
\end{minipage}%
}
\caption{Illustrations of the AV's behavior attained by the proposed integrated intention prediction and decision-making framework in four scenarios. Four representative snapshots for each scenario during the performance evaluation are presented with Straight Road ((a)-(d)), Intersection-v0 ((e)-(h)), Intersection-v1 ((i)-(l)), and Roundabout ((m)-(p)), respectively. The green car and blue cars represent the AV and SVs under normal driving conditions, respectively. The color of the trajectory reflects the speed of the vehicle, with the transition from purple to yellow indicating an increase in vehicle speed.}
\label{behavior}
\end{figure*}
To verify the effectiveness of the proposed algorithm, we implement our framework and other baseline methods in the aforementioned scenarios and obtain the learning curves shown in Fig. \ref{scenrios}. It can be observed that the training results of the proposed algorithm are significantly better than those of all baseline methods. 
This could be attributed to the effectiveness of the intention prediction module within the proposed integrated framework. The decision-making process of SVs takes into account both STD and LTO intention, where LTO intention indicates the willingness to complete tasks of SVs and STD reflects short-term obstacle avoidance intention. The AV predicts the intention of SVs, thereby making decisions accordingly in quick succession, resulting in greater rewards than other baseline methods.
Compared with the ablation study of the PPO method without these intention predictions, the method we proposed achieves a reward increase in all scenarios, but just a little improvement in the roundabout scenario. The reason is that our framework has some difficulties in taking spatio-spectrum into account, which significantly affects the performance in the roundabout scenario, due to its complex road structure. When compared with two other baseline methods, the improvement is even greater.

To further evaluate the performance of the integrated intention prediction and decision-making framework, we evaluate the trained policy in the test environment and set three metrics to assess the effectiveness of each algorithm: (1) we define the success rate as the ratio of successful task completions in 100 trials to reflect the ability of different algorithm-trained policies to complete the task; (2) we define the efficiency of the trained policy as $ \left({t_{\text {max}}- t} \right)/ \left({t_{\text {max }}-t_{\text {min }}}  \right)$, where $t_{\text {max}}$ and $t_{\text {min}}$ are the time thresholds for task completion;  (3) we define the safety as the proportion of the AV that does not collide with SVs or the road boundary in 100 trials. The results are shown in Table \ref{table_metrics}. It can be observed that the framework we proposed achieves the best performance in all four scenarios, which can be attributed to predicting the LTO and STD intention of SVs, enabling the agent to strike a balance between safety and efficiency, thus yielding better results.

\begin{table*}[]
\centering
\caption{Comparison of success rate, efficiency, and safety among different methods in various driving scenarios.
}
\label{table_metrics}
\begin{tabular}{c|ccc|ccc|ccc|ccc}
\hline
\multirow{2}{*}{Methods} & \multicolumn{3}{c|}{Straight Road}            & \multicolumn{3}{c|}{Intersection-v0}         & \multicolumn{3}{c|}{Intersection-v1}             & \multicolumn{3}{c}{Roundabout}    \\ \cline{2-13} 
                          & succ.(\%) & eff.(\%) & safety(\%) & succ.(\%) & eff.(\%) & safety(\%) & succ.(\%) & eff.(\%) & safety(\%) & succ.(\%) & eff.(\%) & safety(\%)  \\ \hline
A2C                     &65 &74 &79 &70 &92 &72 &42 &91 &46 &67 &54 &81   \\
PPO                     &75 &70 &81  &68 &88 &71 &67 &81 &68 &91 &82 &94 \\
DQN                     &61 &67 &77 &68 &84 &73 &66 &71 &73 &61 &88 &65\\
Ours                 &\textbf{93} &\textbf{86} &\textbf{95} &\textbf{ 88} &\textbf{92} &\textbf{91} &\textbf{87} &\textbf{90} &\textbf{89} &\textbf{93} &\textbf{92} &\textbf{96}  \\ \hline
\end{tabular}
\end{table*}

Fig. \ref{behavior} shows the interactive behavior of the AV in the four scenarios. As a representative example, in the scenario of Intersection-v1, the goal of the AV is to safely and efficiently navigate the intersection. In the left turn task, the AV notices the intention of a vehicle making a left turn and moving slowly. So the AV chooses to change lanes before a conflict occurs to avoid a collision. When the AV detects a lane conflict with the planned route of the same vehicle in front, it opts to change lanes to a safer but further lane. This behavior demonstrates the trade-off considered by the AV between efficiency and safety, and the AV selects a safer decision after evaluation. Here, we only visualize the AV's trajectory in the representative snapshots. 

\section{CONCLUSIONS}

 
 In this paper, we propose a novel integrated intention prediction and decision-making framework for autonomous driving systems from the perspective of the frequency domain. It is worth noting that a spectrum attention net is designed to capture the trend of each frequency component over time and their interrelations, so that the intention of SVs can be predicted. Besides,
 the PPO algorithm tackles the non-stationary problem in the integrated framework by employing a clipping mechanism within its objective function.
Experiments in four representative traffic scenarios are conducted to verify the effectiveness of the proposed framework.
The results show that the proposed framework outperforms baselines across various metrics in terms of the success rate of completing driving tasks, efficiency, and safety.
Future work involves taking advantage of frequency domain representation in dealing with uncertainty in autonomous driving tasks.

\addtolength{\textheight}{-6cm}   









\bibliographystyle{ieeetr}
\bibliography{root}
\end{document}